# Separated Attention: An Improved Cycle GAN Based Under Water Image Enhancement Method

Tashmoy Ghosh
*School of Electronics Engineering*
*Vellore Institute of Technology, Chennai*
Chennai, India
tashmoy.ghosh2020@vitstudent.ac.in

***Abstract*—In this paper we have present an improved *Cycle GAN* based model for under water image enhancement. We have utilized the cycle consistent learning technique of the state-of-the-art *Cycle GAN* model with modification in the loss function in terms of depth-oriented attention which enhance the contrast of the overall image, keeping global content, color, local texture, and style information intact. We trained the *Cycle GAN* model with the modified loss functions on the benchmarked Enhancing Underwater Visual Perception (EUPV) dataset a large dataset including paired and unpaired sets of underwater images (*poor* and *good* quality) taken with seven distinct cameras in a range of visibility situation during research on ocean exploration and human-robot cooperation. In addition, we perform qualitative and quantitative evaluation which supports the given technique applied and provided a better contrast enhancement model of underwater imagery. More significantly, the upgraded images provide better results from conventional models and further for under water navigation, pose estimation, saliency prediction, object detection and tracking. The results validate the appropriateness of the model for autonomous underwater vehicles (AUV) in visual navigation.**

## I. Introduction

Autonomous Underwater Vehicles (AUV) and Remotely Teleoperated Vehicle (RTV) became the spot of widely used vehicles for under water exploration such as the monitoring of marine ecosystem such as species migration and coral reefs [1], inspection of bridge structures [2], pipelines and wreckage. Underwater mine detection [3], Reconstructing underwater dynamic scene analysis, seabed mapping, human-robot collaboration [4], and many more. But the perception system suffers the major challenges of underwater visual conditions [5]. Despite using high-quality cameras visuals are affected by low visibility, poor contrast, light refraction, absorption, and scattering [6]. As the propagation of light differs underwater, supported by scattering and varying underwater optical properties thus producing strange set of nonlinear distortions and colour degraded images. The visual features are greatly attenuated with blue or greenish hue, as the red wavelength are absorbed as light travels further in deep underwater. This blurry visual affects the contrast between the foreground and background features [7] and as a result the performance of vision-based task such as object tracking, detection, classification, segmentation and visual servoing for autonomous navigation [8] are degraded significantly. A robust and accurate image enhancement method is required which has the capability to handle differential features over dynamic underwater visual scenarios thereby restoring the perceptual and statistical quantities efficiently. General physics-based model [9] are used for dehazing and colour correction purpose but they lack generalization and produces uneven artifacts. These models are constrained towards processing specific set of features and requires frequent adjustment in their statistical parameters to get the desired output. Besides these models are computational expensive for real time processing.

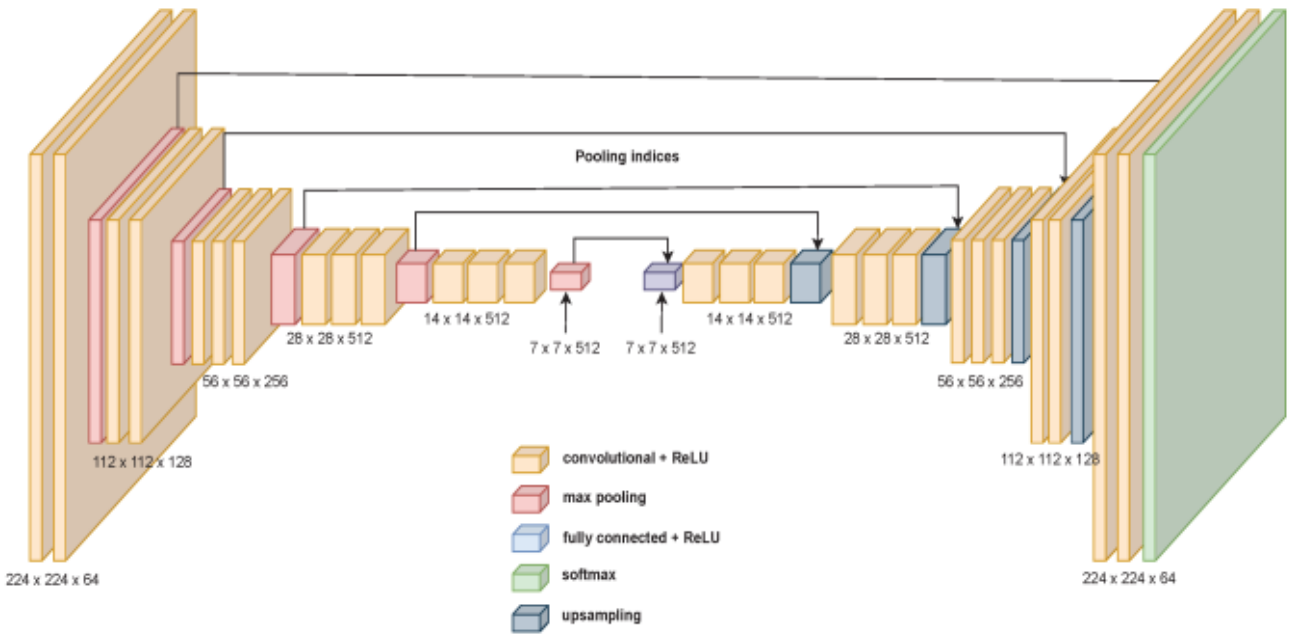

Fig 1. VGG16 based Autoencoder [11]

Recently learning based methods have showed significant improvement in image processing tasks, thanks to its level wise features memorising capabilities from low to mid and then high-level features. Convolutional Neural Network (CNN) [10] serves as a fundamental model for extracting features over various levels supported by Dense Multi-Layer Perceptron (MLP) [16] for learning perceptual image quality for large collection of diverse paired and unpaired datasets. CNN paved the way for development of various state of the art deep learning model like Variational Autoencoders (VAE) [11] and Generative Adversarial Networks (GAN) which used residual VAE's (as shown in Fig 1) and are utilised for capturing latent features in terms of statistical distributions (mean and variance) in lower dimensional space. The distributions can be altered to generate artificial features and easily reconstructed back to same dimensions with modified image quality. Style Transfer [12] is one of the recent advancements in image-to-image translation and transformation which utilises GAN models for automatic colour enhancement, dehazing, super resolution [13] without affecting much the original image quality. Still there is significant space for improvements as there is a possibility of unknown variability in data space which the model has never mapped. Nevertheless, underwater research serves to be a significant field for testing this enhancement model considering the underlying problem for poor perceptual vision.

We try to address the challenges of contrast enhancement in underwater imagery by analysing the applicability and feasibility for real time underwater perception. We utilised the *Style Transfer* as one of the image-to-image translation method where the model takes an unenhanced (or distorted) image and learns to map to its corresponding enhanced version. We modified the loss parameters of the *Cycle GAN* [14] model and allowed it to be trained on the benchmarked large scale EUPV (Enhancement of Underwater Visual Perception) dataset [15] as shown in Fig .2.

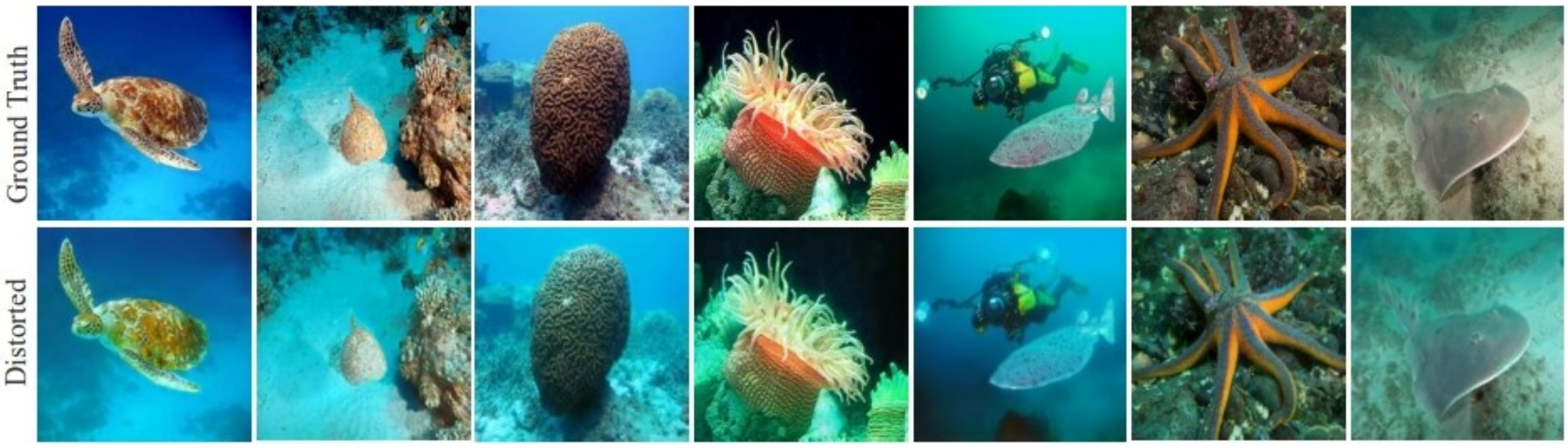

Fig 2. EUPV Dataset [15] with shown Distorted and Ground Truth Images

The following are the contribution in this paper:

a) We present a improved *Cycle GAN* based model for underwater image enhancement with the capability of seperated learning by modifying its loss function which can help in defining usefull contrast in the enhanced image for better perception. Consecutively we introduced attention coefficients for getting adaptive attention between foreground and background.
b) We utilised depth map as an attention map for feature separation to learn the progressive depth-oriented features at the last convoluted layers of the generator.
c) Furthermore, we provide a qualitative and quantitative performance evaluation compared to the start of the art models. We even tested our model for object detection purpose, for an underwater scene.

In addition, we even performed a user study for evaluating the effectiveness of model performance over varying test condition and got impressive results.

## II. RELATED WORK

### A. *Image Enhancement*

Image enhancement is one of the pre-processing steps in the area of computer vision, signal processing and robotics. Conventional approaches [9] use manually parametrised filters to maintain brightness, contrast and colour consistency. Histograms are utilised for statistical analysis of the scene and based on that global enhancement, de-blurring and de-hazing task are acquired. But with the advent of the deep learning methods (especially CNN [10] based model), single image enhancement process is well developed for mapping improved colourisation, contrast, de-blurring and hazing process [18] with the capability of learning features over a large-scale dataset. CNNs based models have several levels of features extraction in terms of features maps using non-linear symmetric filters from low level features in terms of object boundaries (sharpness), encoded colours (intensities) to high level features in terms of object position, orientation and occupancy in saliency map. This multi-level features collectors with Multi-Layer Perceptron (MLP) [16] learns to maps multiple images to its desired output and preforms much better results in an unsupervised fashion compared to hand-crafted filters when tested with new image data.

GAN models which are crafted from the CNN based models like *Autoencoders,* [17] *(as shown in Fig 1.)* are recently catching attention in *Style Transfer* process for image-to-image mapping problems. This type of models has two essential entities named as *generator* and *discriminator* which involves in a min-mix game where the *generator* tries to fool the *discriminator* by generating fake images which appears to be sampled from the combination of the real image with some gaussian noise (to sharpen image and avoid mode collapse). Consecutively the discriminator tries to differentiate between the real and fake image penalise the generator for producing fake image and also penalising itself for not detecting correctly the fake image in terms of loss. Eventually the discriminator gets better in discarding the fake image and generator learns the image distribution consecutively thereby reaching an equilibrium point (minimizing the individual loss).

The *Loss* functions becomes the modifier unit for varying the training process. There are several GAN models which focus modifying their loss functions like Energy based GAN [19] used Lyapunov functions for discriminator stability, Least-Square GAN [20] uses least square based loss function for addressing the vanishing gradient problem, Wasserstein GAN [21] uses Earth Mover Distance measuring difference between the model and data distribution for modelling it loss function. On the other hand, Conditional GAN [22] put constraint on the generated to data which follows a patterns or distribution and similarly the PIX2Pix [23] model used these conditional parameters to map between an arbitrary input domain (distorted or unenhanced image) and the required output domain (enhanced image).

The fact that the aforementioned models depend on paired training data, which is difficult or impossible to get for many real-world applications, is one of their main drawbacks. This issue is resolved by the two-way GANs (such as Cycle GAN [14], Dual GAN [24], etc.) which use a "cycle-consistency loss" to enable the learning of mutual mappings between two domains using unpaired data. Additionally, unpaired learning of perceptual image enhancement has been successfully accomplished with such models [25]. Moreover, the quality of picture augmentation using GANs is enhanced by the insertion of loss-terms for maintaining the high-level feature-based information [26].

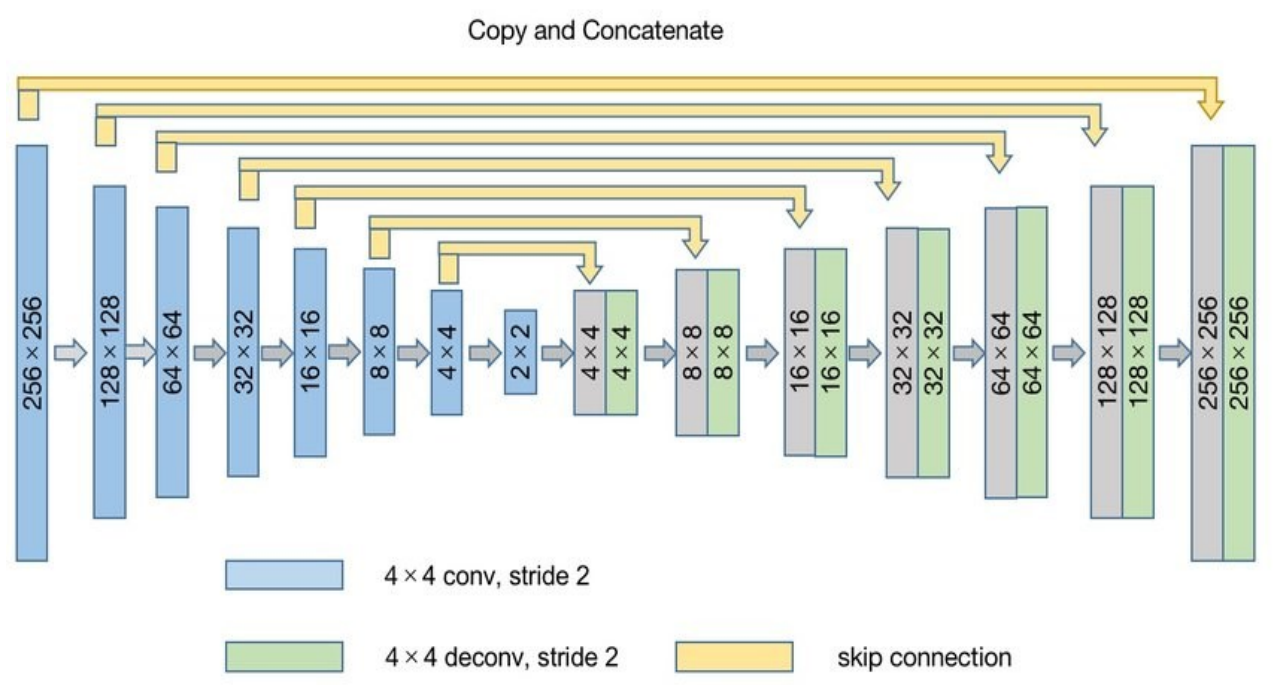

Fig 3. Generator: U Net [32] Architecture

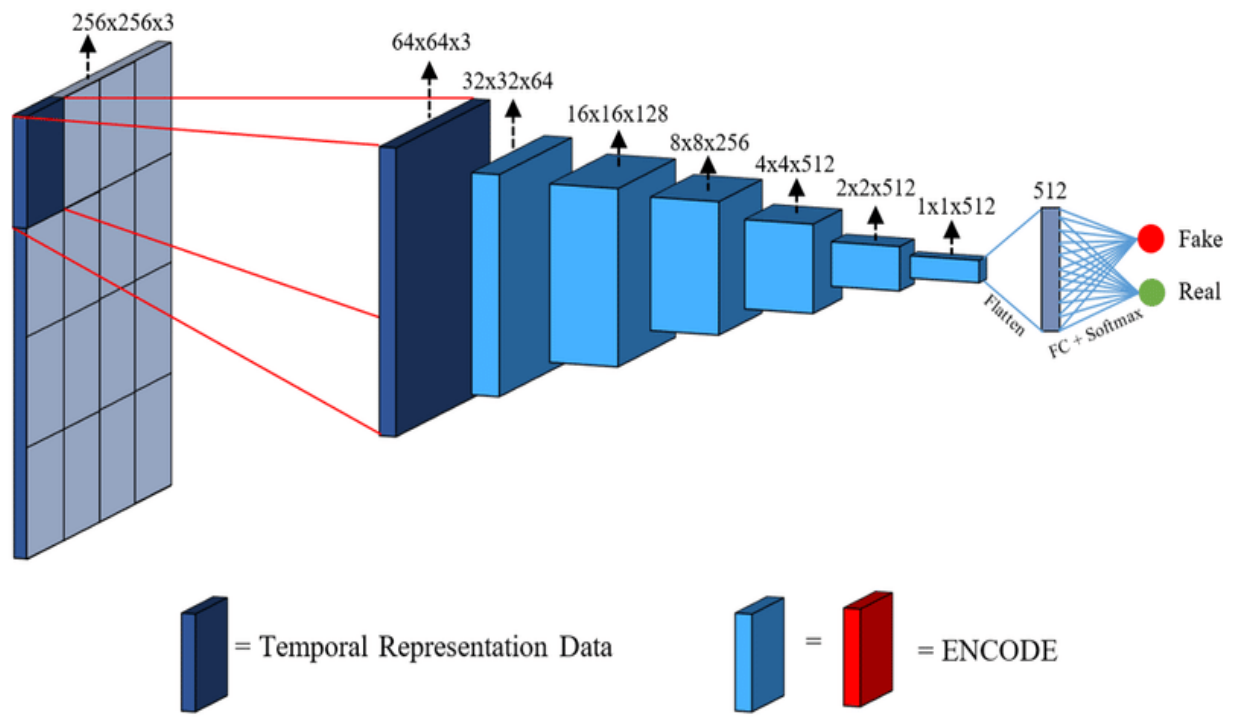

Fig 4. Discriminator: Markovian Patch GAN [33] Architecture

### B. *Enhancing Underwater Visual Perception*

Conventional physics-based methods use Light Spectrum optimization [27] by using light source for surrounding which can penetrate water easily and simultaneously optimise the selection of wavelength. Polarised light source [28] also helps in reducing glare and increase contrast. Dehazing algorithms reduce scattering by estimating the transmission map of the scene and performs restoration-based transmission intensity. The newest Light Field Enhancement [29] approaches, model the propagation of light underwater using principles of optics and fluid dynamics to simulate and enhance the light field in underwater scenes. These methods may involve simulating the interactions of light with water particles and surfaces to generate more realistic renderings of underwater environments. Adaptive optics systems [30] dynamically correct for distortions caused by the turbulent nature of water, which can degrade image quality.

Besides this the single image enhancement models based on adversarial networks which the residual learning capabilities have proved to be the recent advancements. The are known for creating enhanced synthetic scenes from the distorted visual data being trained on large scale paired or unpaired underwater visual improvement dataset. But for paired training, they usually only employ artificially distorted pictures, which often reduces their generalisation performance especially in terms of colour and contrast oriented features due to lack of saliency information. And it limits their applicability for improving real-time visual perception. So, an attempt is made to address these aspects in this paper.

## III. PROPOSED IMPROVED MODEL AND DATASET

We utilised a *Cycle GAN* based model and introduced a new method of training this based on separated attention (Fig 6). We observed that images with similar and plentiful foreground and background features are difficult to reconstruct by keeping the contrast features intact. This happens due to saturated intensity gradients of the feature maps. So, we planned to extracted the foreground and background and trained our *Cycle GAN* model parallelly for mapping both the differentiated features but separately.

### A. *Cycle GAN Architecture*

Provided with a source domain image $X$ (distorted or unenhanced image), our motive is to learn a nonlinear mapping G: X to Y with Y as enhanced output image. Simultaneously a backward mapping F: Y to X is also preformed to get the source domain image which is mentioned a *cycle consistency* [31]. This bi-directional mapping process captures special characteristics of one image collection and figure out how these characteristics could be translated into the other image collection, all in the absence of any paired training examples. The model learns to adapt mapping domains with unpaired input-output image sets there for provides an unsupervised training process. The generator network for X to Y and Y to X mapping is designed using a U-Net structure [32] (Fig 3). Basically, it has an encoder and decoder units ($e_1$-$e_7$) which tries to perform dimensionality reduction of the feature space with successive convolution and the other one ($d_1$-$d_7$) tries to reconstruct the feature map back to source image. The encoder and decoder units have residual connections between the mirrored layers, i.e., between ($e_1$, $d_7$), ($e_2$, $d_6$), ($e_3$, $d_5$), ($e_4$, $d_4$), ($e_5$, $d_3$), ($e_6$, $d_2$) and ($e_7$, $d_1$). The idea of skip connection tries to get the contextual features as feedback from the encoder layer which helps to reconstruct the source image better. The network takes $256 \times 256 \times 3$ image and passed it to 2D convolutional layer with $4 \times 4$ filter followed by a Leaky-Re-LU non-linear function and Batch Normalization (BN) to produce 256 features map of size $8 \times 8$ at the encoder unit. Correspondingly decoder utilizes these feature-maps and inputs from the skip-connections to learn to generate a $256 \times 256 \times 3$ (enhanced) images as output.

We use a Markovian Patch GAN [33] (Fig 4) architecture for the discriminator, which only makes judgements based on patch-level information and assumes the independence of pixels beyond the patch-size. It is crucial to make this assumption in order to properly collect high-frequency information such regional style and texture. Furthermore, compared to universally discriminating at the picture level, this setup uses fewer parameters, making it computationally more efficient. A $256 \times 256 \times 6$ input (a real and produced picture) is transformed into a $64 \times 64 \times 1$ output, which is the discriminator's averaged, and downscaled further below using five convolutional layers. Using a stride of 2, $3 \times 3$ convolutional filters are applied at each layer, followed by the application of BN and non-linearity Leaky Re-LU function in the same manner as the generator.

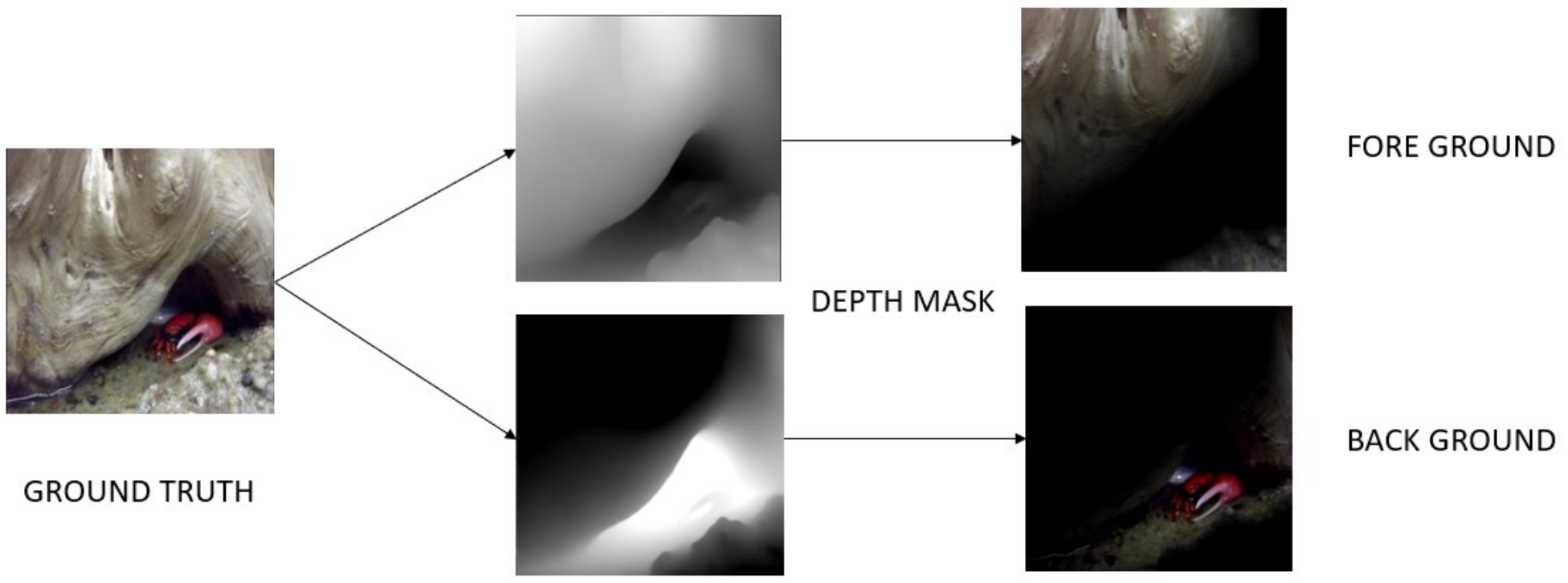

Fig 5. Feature Separation using masking with Depth Map (generated from Depth Anything Model).

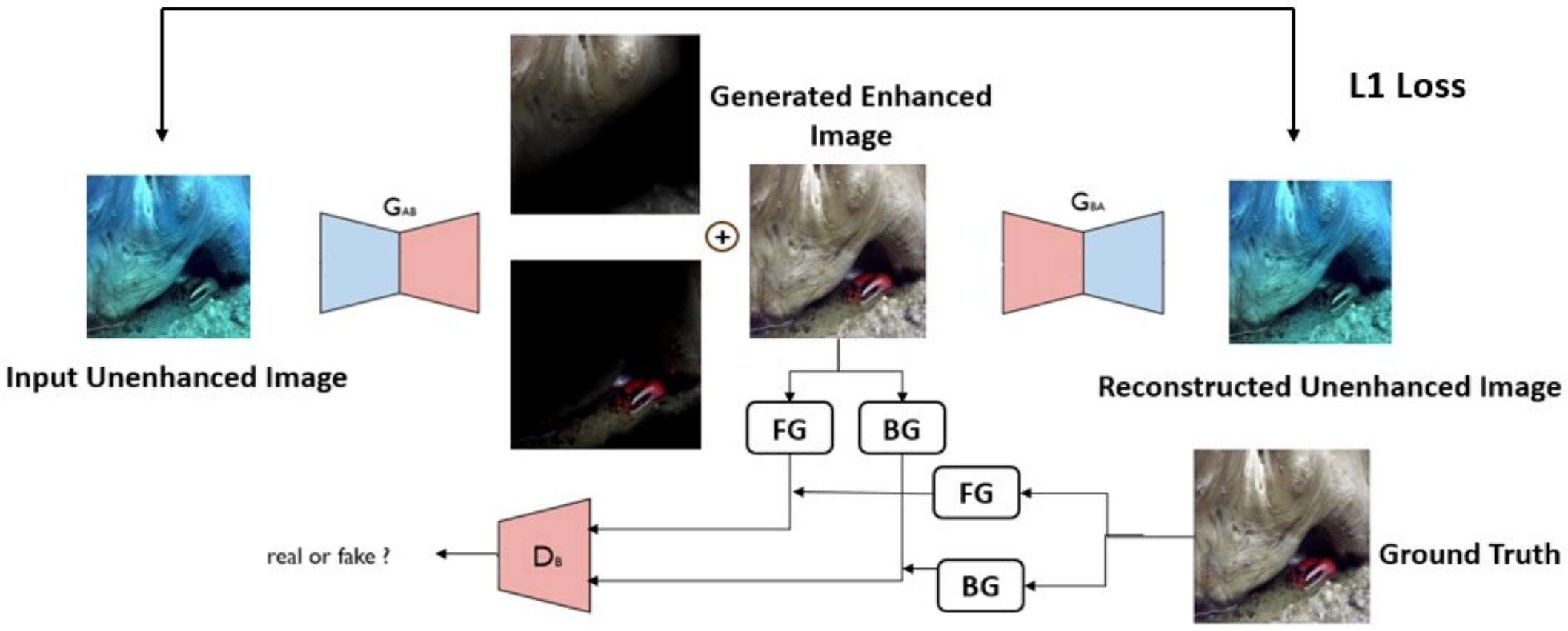

Fig 6. Process Flow of the modified Cycle GAN Architecture

### B. *Depth Attention*

We utilized depth map [34] of the enhanced ground truth domain image (Fig 5) which is created from the *Depth Anything* [35] which, stabilized Monocular Depth Estimation (MDE) foundation model with impressive features as zero-shot depth estimation, zero-shot metric depth estimation and domain fine tuning. Depth Anything is trained on 1.5M labeled images and 62M+ unlabeled images and evaluated on the benchmarked NYUv2 and KITTI dataset. The *Depth Anything* model produces our desired grayscale depth map of enhanced ground truth images of the *EUPV* dataset [15]. The model helps in producing rich semantic information with sharp change in intensity gradient thereby informing about the change in the relative depth-oriented features of the image and it is finally used for features separation as described in the following section.

### C. *Features Seperation*

Considering the prime objective of our approach we try to extract the foreground and background features of the generated as well as the ground truth enhanced images using the depth attention map as described above. We take a 256 × 256 × 3 input image matrix **I** and we normalize its intensity value over the range of 0-255, whereas the depth map matrix **D** of same dimensions as that of the input with its intensity normalized between 0-1. Finally, the depth map is masked (Fig 5) as shown in equation 1 with input image in terms of elementwise multiplication of the pixel intensity over the three channels.

$$Mask\ (I, D)\ =\ I\ .*\ D \qquad (1)$$

So, the pixels with low intensity value in the depth map, which are close to 0 will suppress the corresponding areas of the input image whereas the areas with depth intensity close to 1 will keep the corresponding features in the input image intact.

We utilized this masking technique to separate our foreground and background features respectively (Fig 5). For foreground features we normally use our produced depth map masking with the input image as mentioned in equation 2, whereas for background features, we invert out depth map by subtracting the depth map pixel intensities by 1 and finally mask the inverse depth map with the input image as mentioned in equation 3.

*Foreground Features:*

$$Mask\ (I, D)\ =\ I\ .*\ D \qquad (2)$$

*Background Features:*

$$Mask\ (I, ID)\ =\ I\ .*\ ID \quad (3)$$

Where $ID\ =\ 1 - D$

## D. Loss Function Formulation

Our goal is to learn objective mapping between X and Y domain with associated training samples $\{x_i\}_{i=0}^{N}$ where $x_i$ € X and $\{y_i\}_{i=0}^{N}$ where $y_i$ € Y. The data is presented in terms of probability distribution as x ~ $p(x)$ and y ~ $p(y)$. The model includes two mappings G: X to Y and F: Y to X. In addition, two adversarial discriminators $D_X$ and $D_Y$ were introduced, where DX aims to differentiate between images {x} and translated images {F(y)}; in the same way, DY aims to discriminate between {y} and {G(x)}. Our objective contains two types of loss terms almost similar to *Cycle GAN*: *adversarial losses* [36] for matching the distribution of generated to the data distribution in the target domain; and *cycle consistency losses* [31] to prevent the learned mappings G and F from contradicting each other. But we introduced the above loss functions for separately mapping the foreground and background features for discriminators and a collective foreground and background loss for generators.

### 1) Adversarial Loss

It provides feedback for mapping function G: X to Y and its discriminator $D_Y$. The following objective function is:

$$\mathcal{L}_{\text{GAN}}(G, D_Y, X, Y) = \mathbb{E}_{y\sim p_{\text{data}}(y)}[\log D_Y(y)] + \mathbb{E}_{x\sim p_{\text{data}}(x)}[\log(1 - D_Y(G(x))], \quad (4)$$

The generator G tries to create an image G(x) similar to Y, while the $D_Y$ job is to distinguish G(x) and real sample y. G tries to minimize the objective function by learning to create nearly similar image to Y, whereas D tries to maximize the loss. A similar objective function is defined for mapping F: Y to X with $D_X$ as the discriminator $\mathcal{L}_{\text{GAN}}(F, D_X, Y, X)$.

### 2) Cycle Consistency Loss

Adversarial loss alone cannot alone cannot guarantee that the learned function can map an individual input x to a desired output y because multiple set of input images can lead to a desired output images considering the network is constrained towards learning a similar distribution of output image. To handle this consistency simultaneous generation as well as restoration is required. For each image x from domain X, the image translation cycle should be able to bring x back to the original image, i.e., x → G(x) → F(G(x)) ≈ x. this is called as forward cycle consistency. Similarly, for each image y from domain Y, G and F should also satisfy backward cycle consistency: y → F(y) → G(F(y)) ≈ y. The desired loss function is defined as:

$$\mathcal{L}_{\text{cyc}}(G, F) = \mathbb{E}_{x\sim p_{\text{data}}(x)}[\|F(G(x)) - x\|_1] + \mathbb{E}_{y\sim p_{\text{data}}(y)}[\|G(F(y)) - y\|_1]. \quad (5)$$

Note the loss mentioned is a L1 loss.

### 3) Final Attention Loss

The Collective objective function is:

$$\mathcal{L}(G, F, D_X, D_Y) = \mathcal{L}_{\text{GAN}}(G, D_Y, X, Y) + \mathcal{L}_{\text{GAN}}(F, D_X, Y, X) + \lambda\mathcal{L}_{\text{cyc}}(G, F), \quad (6)$$

Where λ controls the importance of the two objectives. The total loss is collectively modeled for both foreground and background features (Fig 6) as:

$$L_{attn}(G,F,D_x,D_y) = \mu * L(G_{foreground}, F_{foreground}, D_{X(foreground)}, D_{Y(foreground)}) + \alpha * L(G_{background}, F_{background}, D_{X(background)}, D_{Y(background)}) \quad (7)$$

Where:

$G_{foreground}$, $F_{foreground}$ signifies exclusively foreground features of the generated and restored image after masking.

$D_{X(foreground)}$, $D_{Y(foreground)}$ signifies input to the discriminators as exclusive foreground, ground truth image y and $G_{foreground}$ correspondingly foreground real image x and $F_{foreground}$.

$G_{background}$, $F_{background}$, $D_{X(background)}$, $D_{Y(background)}$ is defined in the similar fashion for background images too.

The μ and α are foreground and background attention parameters which holds value between 1 and 10. This parameter controls the amount to attention required towards enhancement.

So, our final objective is to solve:

*Generators:*

$$G^*, F^* = \arg\min_{G,F} \max_{Dx,Dy} L_{attn}(G, F, D_X, D_Y) \quad (8)$$

*Discriminators:*

$D_X$ and $D_Y$ are trained based on the combined adversarial loss for ground truth and generated images as described in equation (4). But the combined adversarial loss is calculated separately for both foreground and background images and trained parallelly. The objective is to minimize the below losses.

$$L_{GAN}\left(G, D_{Y(foreground)}, X_{(foreground)}, Y_{(foreground)}\right) \quad (9)$$

$$L_{GAN}(G, D_{Y(background)}, X_{(background)}, Y_{(background)}) \quad (10)$$

Similarly, for $D_Y$ separated loss $L_{GAN}(G, D_X, Y, X)$ is calculated and parallelly trained.

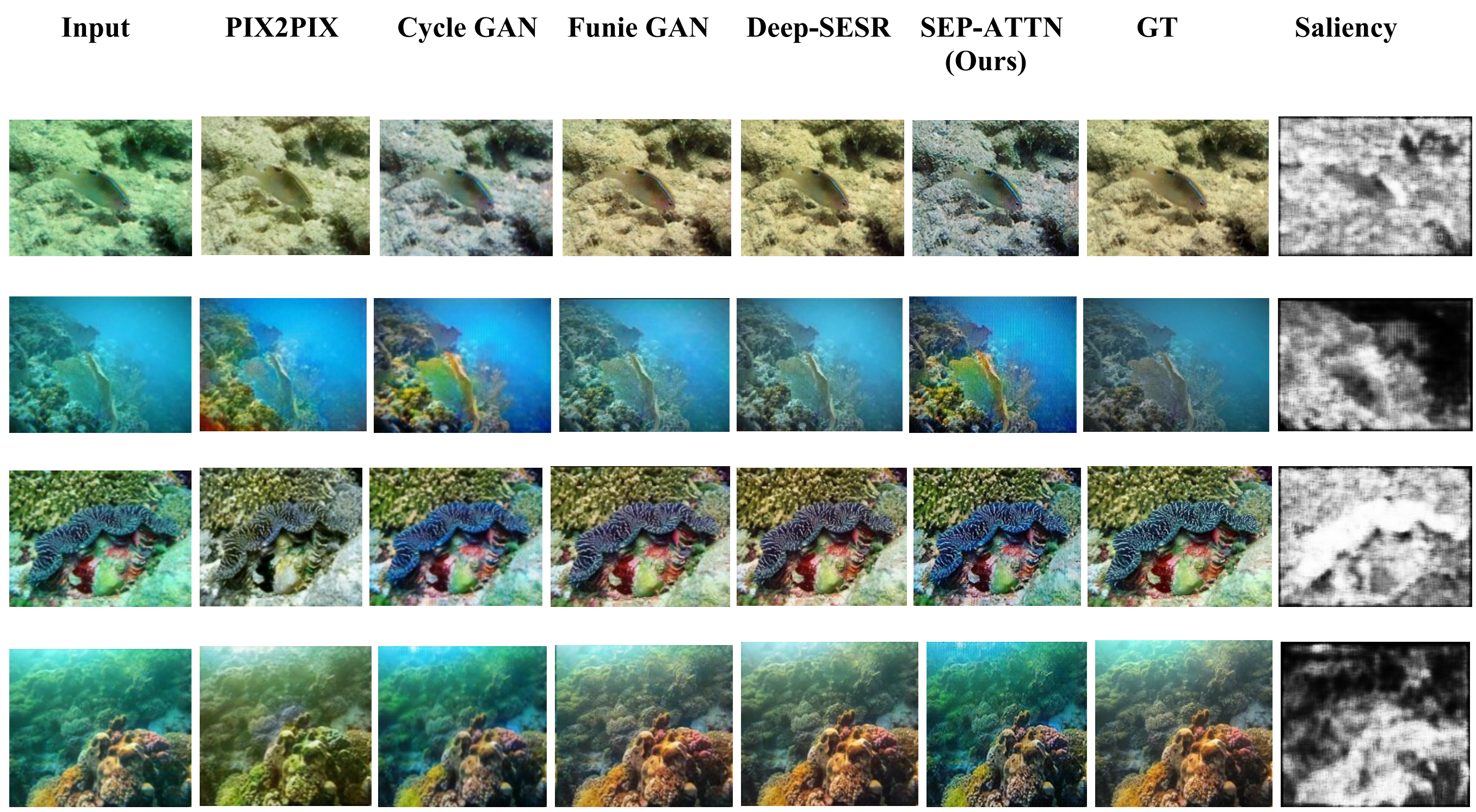

Fig 7. Comparisons of Qualitative performance of our Sep-Attn GAN model with (i) PIX2PIX (ii) Cycle-GAN (iii) FUnie GAN (iv) Deep-SESR (*better glare prevent view at 180% zoom*)

### *E. EUPV Dataset*

An extensive collection of paired and unpaired underwater photos with varying levels of perceptual quality may be found in the EUVP dataset. To get the photos for the dataset, it used seven distinct cameras: many Go-Pros [37], the u-Eye cameras from Aqua AUV [38], low-light USB cameras [40], and the HD camera from Trident ROV [39]. The information was gathered during cooperative human-robot tests and maritime expeditions in diverse areas with varying visibility conditions. The collection also contains photos taken from a few publicly accessible YouTube-TM films. A broad variety of natural diversity (sceneries, waterbody kinds, lighting conditions, etc.) in the data is accommodated by the carefully chosen photographs. Six human volunteers visually assess the photos to separate those of good and bad quality, resulting in the preparation of the unpaired data. They examined a number of image characteristics, such as color, contrast, and sharpness, and they thought about whether the scene could be visually understood by looking at the foreground and individual objects. Therefore, the modelling of human perceptual preferences for underwater image quality is supported by the unpaired training. Conversely, the paired data is created by using the method recommended in [41]. In particular, the unpaired data is used to train a Cycle-GAN [14]-based model, which identifies the domain transformation between the high- and low-quality photos. The learnt model then distorts the high-quality photographs to produce the appropriate pairs; we additionally enhance a collection of underwater photos from Flickr and the ImageNet dataset [42].

The EUVP dataset contains about 12K paired and 8K unpaired cases; shows a few samples. It should be emphasized that the goal is not to replicate the underwater optical deterioration process for image restoration, which necessitates scene depth and waterbody parameters, but rather to permit perceptual image improvement to improve robotic scene interpretation.

### *F. Training Process*

A paired training is performed with the respective loss function as described in equation 8 which guides the G to learn to improve the perceptual image quality so that the generated image is close to the respective ground truth in terms of its global appearance and high-level feature representation. On the other hand, D will discard a generated image that has locally inconsistent texture and style. In particular for GAN loss the negative log likelihood loss is replaced with least square loss for more stability as described in the Cycle GAN paper [14]. So G is trained to minimize $\mathbb{E}_{x\sim p_{\text{data}}(x)}[(D(G(x))-1)^2]$ and D tries to minimize $\mathbb{E}_{y\sim p_{\text{data}}(y)}[(D(y)-1)^2] + \mathbb{E}_{x\sim p_{\text{data}}(x)}[D(G(x))^2]$.

Considering the $\lambda$ parameter we kept it as 10 similar to Cycle GAN [14] paper and kept the $\mu$ and $\alpha$ parameter to be 7 and 3 respectively to maintain a proper contrast. A batch size of 5 is maintained and all the networks were trained with 0.0002 learning rate for 100 epochs.

## IV. EXPERIMENTAL RESULT

We utilised Pytorch libraries [44] to implement the Seperated Attention GAN model. It is trained on 11K paired image data and the rest are used for respective validation and testing. We used single NVIDIA$^{TM}$ GeForce RTX 3080 graphics cards are used for processing and the models are trained for 60K-70K iterations with a batch-size of 5. We now present the experimental evaluations based on a qualitative analysis, standard quantitative metrics, and a user study.

### A. *Qualitative Evaluation*

Qualitatively we observed that our model is able to enhance the colour, contrast with respect to the ground truth images (Fig 7). The greenish blue hue is removed to a greater extent thereby producing a proper colour contrast between foreground and background. The global contrast is equally improved. Therefore the problem of saturated intensity gradient is improved to a greater extent. We observed that the L1 loss helps in capturing fine textures whereas our separated attention loss functions helps in distinguishing colour contrast between background and foreground. We even extracted the last convolutional layers of the decoder to get the reconstructed feature map and a good difference in contrast features are identified.

Successively, we compared our generated output with three state of the art GAN model: (i) PIX2PIX [43] (ii) FUnie-GAN [45] (iii) Cycle-GAN [14] (iv) Deep-SESR [46] and got good results in terms of contrast (as shown in Fig 7).

### B. *Quantitative Evaluation*

We considered two essential metrics for evaluation our model i.e. Peak Signal-to-Noise Ratio (PSNR) and Structural Similarity (SSIM) [47]. We compared quantitatively between the enhanced images produced by our model with the ground truth images. PSNR describes the reconstruction quality of the generated images by comparing it with the ground truth labels using Mean Squared Error (MSE) value.

$$PSNR(\mathbf{x},\mathbf{y}) = 10\log_{10}\left[255^2/MSE(\mathbf{x},\mathbf{y})\right] \quad (11)$$

Whereas SSIM distinguishes the image patches based on three characteristics: luminance, contrast, and structure.

$$SSIM(\mathbf{x},\mathbf{y}) = \left(\frac{2\mu_{\mathbf{x}}\mu_{\mathbf{y}}+c_1}{\mu_{\mathbf{x}}^2+\mu_{\mathbf{y}}^2+c_1}\right)\left(\frac{2\sigma_{\mathbf{xy}}+c_2}{\sigma_{\mathbf{x}}^2+\sigma_{\mathbf{y}}^2+c_2}\right) \quad (12)$$

In Eq. 12, $\mu_X$ ($\mu_Y$) denotes the mean, and $\sigma_X^2$ ($\sigma_Y^2$) is the variance of x (y); whereas $\sigma_{XY}$ denotes the cross correlation between x and y. Additionally, $c_1 = (255 \times 0.01)^2$ and $c_2 = (255 \times 0.03)^2$ are constants that ensure numeric stability.

TABLE I. QUANTATIVE ANALYSIS OF DIFFERENT MODELS WITH AVERAGE PSNR AND SSIM VALUE OVER 1K TEST IMAGES

| S.NO | COMPARATIVE OVERVIEW | | |
|---|---|---|---|
| | ***MODEL*** | ***PSNR (G(X), Y)*** | ***SSIM (G(X), Y)*** |
| 1. | Input | 19.96 ± 2.06 | 0.694 ± 0.028 |
| 2. | PIX2PIX | 22.51 ± 2.37 | 0.711 ± 0.051 |
| 3. | Cycle GAN | 23.41 ± 2.41 | 0.729 ± 0.033 |
| 4. | FUnIE GAN | 23.49 ± 2.92 | 0.733 ± 0.039 |
| 5. | Deep-SESR | 23.28 ± 3.01 | 0.710 ± 0.042 |
| 6. | SEP-Attn GAN (Ours) | 23.79 ± 2.53 | 0.741 ± 0.046 |

In Table I we provide average PSNR and SSIM value over 1k test images and compared the values with the mentioned models. We were able to perform well in comparison to the other models. A equivalent analysis is also made with r Underwater Image Quality Measure (UIQM) [48], which quantifies underwater image colourfulness, sharpness, and contrast. The results is present in Table II.

TABLE II. QUANTATIVE ANALYSIS OF DIFFERENT MODELS WITH AVERAGE UIQM VALUE OVER 1K TEST IMAGES

| S.NO | *MODEL* | UIQM |
|---|---|---|
| 1. | Input | 2.94 ± 0.369 |
| 2. | PIX2PIX | 3.01 ± 0.301 |
| 3. | Cycle GAN | 3.03 ± 0.306 |
| 4. | FUnIE GAN | 3.05 ± 0.305 |
| 5. | Deep-SESR | 3.01 ± 0.310 |
| 6. | SEP-Attn GAN (Ours) | 3.17 ± 0.302 |

The evaluation results indicates that SEP-Attn GAN, Cycle – GAN , FUnIE GAN produce better results. The L1 loss in all the models contributes to 2.9% improvement in the enhancement whereas separated loss accompanies the L1 with an improvement of 7.25 % so, net 4.35% additional improvement achieved.

### C. *User Study*

A standard user study perception analysis is performed. A total of 66 individuals were involved, they were given with 6 images including unenhanced image and generated enhanced images from all the models and asked to rank the top 3 images. It was observed around 78% of individual selected SEP-Attn GAN, Cycle – GAN, FUnIE GAN as the top 3 images with SEP-Attn GAN scored 48% in rank 1, 62% in rank 2 and 83% in rank 3 whereas Cycle-GAN achieved (23,40,50) % and FUnIE GAN achieved (42,51,72) %. Hence our model enhanced the quality for better perceptibility.

### D. *Visual Perception*

With the contrast enhancement feature achieved by out model it provided better perception capabilities in terms of object detection [49]. The better color contrast features between the foreground and background helps in efficiently detecting the objects [50] (as shown in Fig 8) even with low average confidence threshold between 10 to 30%.

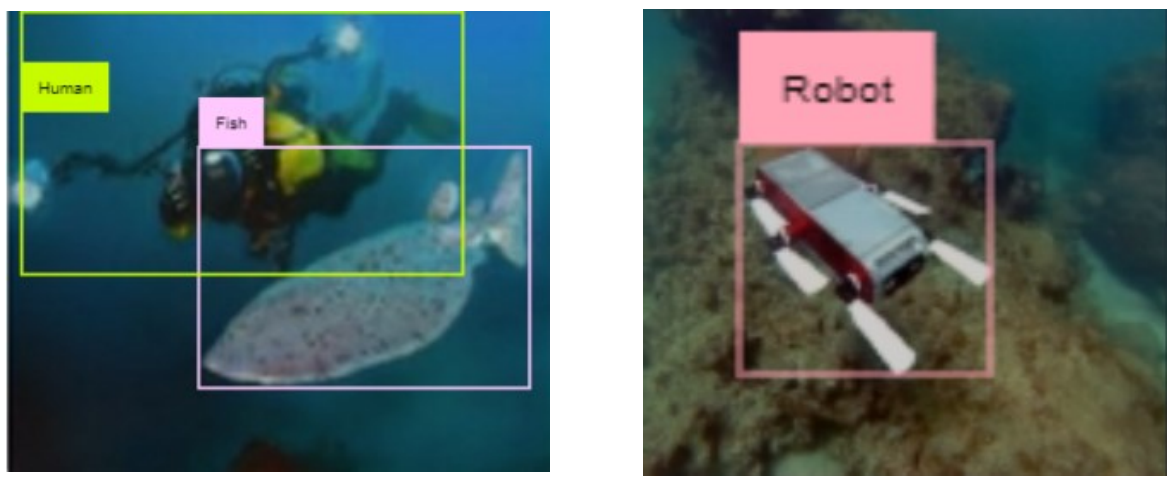

Fig 8. Object Detection using SEP-Attn Model

### *E. Limitations*

With improved efficiency our model suffers from instability too, in few of the challenging cases. First of all, this extensive model takes a bit more time to converge in comparison to standard Cycle GAN model, the reason is that our model used a residual U-Net architecture and also have a extended loss function for the generators and discriminators in terms of foreground and background which increases it processing time by around 36%. Secondly, we noticed a zitter appearing at boundaries of transition between foreground and background. This mostly occurs in images with similar foreground and background textures. The main reason of this problem can the hard constraints of the attention parameters α and μ. These are the problems we would like to address as a part of future work.

## Conclusion

We attempted to improve the contrast enhancement process by separating the foreground and background features for better perception thereby modifying the objectives of the state of the art Cycle GAN model. The separated attention process independently able to maintain the global colour, local content and texture. We trained our model in a supervised fashion, on the benchmarked EUPV dataset which captures various underwater visual information. Consecutively a qualitative and quantitative evaluation is preformed to support the efficiency of our model. The user survey equally helps in distinguishing our models accomplishment over the other start of the art models. Moreover, the paper showcased the efficiency of differential (or separated) contrast enhancement in object detection and saliency prediction task. As part of future work we would like to execute a feasibility test on real time camera feed in underwater environment and check its stability for object detection while maintaining colour consistency, and contrast. Additional we will try to model the attention coefficients α and μ in accordance to the distribution of the features space to get firmness in terms of smooth transitioning between foreground and background features.

## References

[1] Yang, Zhijie, et al. "Exploration and genome mining of natural products from marine Streptomyces." Applied microbiology and biotechnology 104 (2020): 67-76.J. Clerk Maxwell, A Treatise on Electricity and Magnetism, 3rd ed., vol. 2. Oxford: Clarendon, 1892, pp.68–73.

[2] Hou, Shitong, et al. "Underwater inspection of bridge substructures using sonar and deep convolutional network." Advanced Engineering Informatics 52 (2022): 101545.K. Elissa, "Title of paper if known," unpublished.

[3] Munteanu, Dan, et al. "Sea mine detection framework using YOLO, SSD and EfficientDet deep learning models." Sensors 22.23 (2022): 9536.Y. Yorozu, M. Hirano, K. Oka, and Y. Tagawa, "Electron spectroscopy studies on magneto-optical media and plastic substrate interface," IEEE Transl. J. Magn. Japan, vol. 2, pp. 740–741, August 1987 [Digests 9th Annual Conf. Magnetics Japan, p. 301, 1982].

[4] Islam, Md Jahidul, Marc Ho, and Junaed Sattar. "Understanding human motion and gestures for underwater human–robot collaboration." Journal of Field Robotics 36.5 (2019): 851-873.

[5] Huang, Hai, et al. "A review on underwater autonomous environmental perception and target grasp, the challenge of robotic organism capture." Ocean Engineering 195 (2020): 106644.

[6] Zhou, Jingchun, et al. "Underwater image enhancement method with light scattering characteristics." Computers and Electrical Engineering 100 (2022): 107898.

[7] Almutiry, Omar, et al. "Underwater images contrast enhancement and its challenges: a survey." Multimedia Tools and Applications (2021): 1-26.

[8] Lee, Min-Fan Ricky, and Ying-Chu Chen. "Artificial intelligence based object detection and tracking for a small underwater robot." Processes 11.2 (2023): 312.

[9] Raveendran, Smitha, Mukesh D. Patil, and Gajanan K. Birajdar. "Underwater image enhancement: a comprehensive review, recent trends, challenges and applications." Artificial Intelligence Review 54 (2021): 5413-5467.

[10] Ren, Yong, and Xuemin Cheng. "Review of convolutional neural network optimization and training in image processing." Tenth International Symposium on Precision Engineering Measurements and Instrumentation. Vol. 11053. SPIE, 2019.

[11] Yao, Rong, et al. "Unsupervised anomaly detection using variational auto-encoder based feature extraction." 2019 IEEE International Conference on Prognostics and Health Management (ICPHM). IEEE, 2019.

[12] Singh, Akhil, et al. "Neural style transfer: A critical review." IEEE Access 9 (2021): 131583-131613.

[13] Ni, Zhangkai, et al. "Towards unsupervised deep image enhancement with generative adversarial network." IEEE Transactions on Image Processing 29 (2020): 9140-9151.

[14] Dwarkani, Abhinav, et al. "Unpaired Image-to-Image Translation using Cycle Generative Adversarial Networks [J]." International Journal of Engineering and Advanced Technology (IJEAT) 9.6 (2020).

[15] EUPV DATASET: https://irvlab.cs.umn.edu/resources/euvp-dataset.

[16] Kruse, Rudolf, et al. "Multi-layer perceptrons." Computational intelligence: a methodological introduction. Cham: Springer International Publishing, 2022. 53-124.

[17] Yoo, Jaechang, Heesong Eom, and Yong Suk Choi. "Image-to-image translation using a cross-domain auto-encoder and decoder." Applied Sciences 9.22 (2019): 4780.

[18] K. He, J. Sun, and X. Tang. Single Image Haze Removal using Dark Channel Prior. IEEE Transactions on Pattern Analysis and Machine Intelligence, 33(12):2341–2353, 2010.

[19] J. Zhao, M. Mathieu, and Y. LeCun. Energy-based Generative Adversarial Network. 2017.

[20] M. Mirza and S. Osindero. Conditional Generative Adversarial Nets. arXiv preprint arXiv:1411.1784, 2014.

[21] Ren, Jinfu, Yang Liu, and Jiming Liu. "EWGAN: Entropy-based Wasserstein GAN for imbalanced learning." Proceedings of the AAAI conference on artificial intelligence. Vol. 33. No. 01. 2019.

[22] Boroujeni, Sayed Pedram Haeri, and Abolfazl Razi. "Ic-gan: An improved conditional generative adversarial network for rgb-to-ir image translation with applications to forest fire monitoring." Expert Systems with Applications 238 (2024): 121962.

[23] Henry, Joyce, Terry Natalie, and Den Madsen. "Pix2Pix GAN for image-to-image Translation." Research Gate Publication (2021): 1-5.

[24] Han, Junlin, et al. "Dual contrastive learning for unsupervised image-to-image translation." Proceedings of the IEEE/CVF conference on computer vision and pattern recognition. 2021.

[25] Y.-S. Chen, Y.-C. Wang, M.-H. Kao, and Y.-Y. Chuang. Deep Photo Enhancer: Unpaired Learning for Image Enhancement from Photographs with GANs. In IEEE Conference on Computer Vision and Pattern Recognition (CVPR), pages 6306– 6314. IEEE, 2018.

[26] A. Ignatov, N. Kobyshev, R. Timofte, K. Vanhoey, and L. Van Gool. DSLR-quality Photos on Mobile Devices with Deep Convolutional Networks. In IEEE International Conference on Computer Vision (ICCV), pages 3277–3285, 2017.

[27] Kwon, Sung Min, et al. "Environment ‐ adaptable artificial visual perception behaviors using a light ‐ adjustable optoelectronic neuromorphic device array." Advanced Materials 31.52 (2019): 1906433.

[28] Bergkvist, Max. Studies on Polarised Light Spectroscopy. Vol. 1689. Linköping University Electronic Press, 2019.

[29] Wang, Yingqian, et al. "Spatial-angular interaction for light field image super-resolution." Computer Vision–ECCV 2020: 16th European Conference, Glasgow, UK, August 23–28, 2020, Proceedings, Part XXIII 16. Springer International Publishing, 2020.

[30] Hampson, Karen M., et al. "Adaptive optics for high-resolution imaging." Nature Reviews Methods Primers 1.1 (2021): 68.

[31] Zhu, Ruiqi, Tianhong Dai, and Oya Celiktutan. "Cross Domain Policy Transfer with Effect Cycle-Consistency." arXiv preprint arXiv:2403.02018 (2024).

[32] Huang, Huimin, et al. "Unet 3+: A full-scale connected unet for medical image segmentation." ICASSP 2020-2020 IEEE international conference on acoustics, speech and signal processing (ICASSP). IEEE, 2020.

[33] Bera, Sutanu, and Prabir Kumar Biswas. "Noise conscious training of non local neural network powered by self attentive spectral normalized Markovian patch GAN for low dose CT denoising." IEEE Transactions on Medical Imaging 40.12 (2021): 3663-3673.

[34] Piao, Yongri, et al. "Depth-induced multi-scale recurrent attention network for saliency detection." Proceedings of the IEEE/CVF international conference on computer vision. 2019.

[35] Depth Anything: Unleashing the Power of Large-Scale Unlabeled Data, Yang, Lihe and Kang, Bingyi and Huang, Zilong and Xu, Xiaogang and Feng, Jiashi and Zhao, Hengshuang, CVPR,2024.

[36] Chen, Minghao, et al. "Adversarial-learned loss for domain adaptation." Proceedings of the AAAI conference on artificial intelligence. Vol. 34. No. 04. 2020.

[37] GoPro. GoPro Hero 5. https://gopro.com/, 2016. Accessed: 3-15-2019.

[38] G. Dudek, P. Giguere, C. Prahacs, S. Saunderson, J. Sattar, L.-A. Torres-Mendez, Jenkin, et al. Aqua: An Amphibious Autonomous Robot. Computer, 40(1):46–53, 2007.

[39] OpenROV. TRIDENT. https://www.openrov.com/, 2017. Accessed: 3-15-2019.

[40] BlueRobotics. Low-light HD USB Camera. https: //www.bluerobotics.com/, 2016. Accessed: 3-15- 2019.

[41] C. Fabbri, M. J. Islam, and J. Sattar. Enhancing Underwater Imagery using Generative Adversarial Networks. In IEEE International Conference on Robotics and Automation (ICRA), pages 7159–7165. IEEE, 2018.

[42] J. Deng, W. Dong, R. Socher, L.-J. Li, K. Li, and L. FeiFei. ImageNet: A Large-scale Hierarchical Image Database. In Conference on Computer Vision and Pattern Recognition (CVPR), pages 248–255. IEEE, 2009.

[43] Qu, Yanyun, et al. "Enhanced pix2pix dehazing network." Proceedings of the IEEE/CVF conference on computer vision and pattern recognition. 2019.

[44] Imambi, Sagar, Kolla Bhanu Prakash, and G. R. Kanagachidambaresan. "PyTorch." Programming with TensorFlow: Solution for Edge Computing Applications (2021): 87-104.

[45] Islam, Md Jahidul, Youya Xia, and Junaed Sattar. "Fast underwater image enhancement for improved visual perception." IEEE Robotics and Automation Letters 5.2 (2020): 3227-3234.

[46] Islam, Md Jahidul, Peigen Luo, and Junaed Sattar. "Simultaneous enhancement and super-resolution of underwater imagery for improved visual perception." arXiv preprint arXiv:2002.01155 (2020).

[47] Setiadi, De Rosal Igantius Moses. "PSNR vs SSIM: imperceptibility quality assessment for image steganography." Multimedia Tools and Applications 80.6 (2021): 8423-8444.

[48] Sachin, T. M., and G. P. Prerana. "Underwater Image Enhancement Using Color Correction and Fusion Check for updates." VLSI, Communication and Signal Processing: Select Proceedings of the 5th International Conference, VCAS 2022. Vol. 1024. Springer Nature, 2023.

[49] Wei, Jian, et al. "Enhanced object detection with deep convolutional neural networks for advanced driving assistance." IEEE transactions on intelligent transportation systems 21.4 (2019): 1572-1583.

[50] Amit, Yali, Pedro Felzenszwalb, and Ross Girshick. "Object detection." Computer Vision: A Reference Guide. Cham: Springer International Publishing, 2021. 875-883.